\documentclass{article}
\usepackage{amssymb}
\usepackage{amsmath}
\usepackage{times}
\usepackage{graphicx} 
\usepackage{subfigure} 
\usepackage{float}
\usepackage{placeins}
\usepackage{natbib}

\usepackage{algorithm}
\usepackage{algorithmic}

\usepackage{hyperref}
\usepackage{csquotes}

\usepackage[accepted]{icml2016} 

\newcommand{\XX}{\mathcal{X}}
\newcommand{\FF}{\mathcal{F}}

\newcommand{\R}{\mathbb{R}}

\newcommand{\bS}{\mathbf{S}}

\newcommand{\bB}{\mathbf{B}}

\newcommand{\ba}{\mathbf{a}}
\newcommand{\bx}{\mathbf{x}}
\newcommand{\bk}{\mathbf{k}}

\newcommand{\bI}{\mathbf{I}}
\newcommand{\bK}{\mathbf{K}}
\newcommand{\bX}{\mathbf{X}}

\newcommand{\bs}{\mathbf{s}}

\newcommand{\mS}{\mathcal{S}}

\newcommand{\bA}{\mathbf{A}}

\newcommand{\mkc}{{\bf MKC}}
\newcommand{\psd}{{\bf MKC}$_{sdp}$}
\newcommand{\emht}{{\bf MKC}$_{embd(ht)}$}
\newcommand{\emhm}{{\bf MKC}$_{embd(hm)}$}
\newcommand{\mkckernel}{{\bf MKC}$_{app}$}

\newcommand{\knn}{{\bf Knn}}
\newcommand{\wknn}{{\bf wKnn}}
\newcommand{\emb}{{\bf EM}$_{based}$}

\icmltitlerunning{Multi-view Kernel Completion}

\begin{document} 

\twocolumn[
\icmltitle{Multi-view Kernel Completion }

\icmlauthor{Sahely Bhadra}{sahely.bhadra@aalto.fi}
\icmlauthor{Samuel Kaski}{samuel.kaski@aalto.fi}
\icmlauthor{Juho Rousu}{juho.rousu@aalto.fi}
\icmladdress{Helsinki Institute for Information Technology HIIT \\ Department of Computer Science, Aalto University School of Science and Technology, Finland.}

\icmlkeywords{Multi-view learning, missing data, kernel completion}

\vskip 0.3in
]

\begin{abstract} 
In this paper, we introduce the first method that (1) can complete kernel matrices with completely missing rows and columns as opposed to individual missing kernel values, (2) does not require any of the kernels to be complete {\it a priori}, and (3) can tackle non-linear kernels. These aspects are necessary in practical applications such as integrating legacy data sets, learning under sensor failures and learning when measurements are costly for some of the views. The proposed approach predicts missing rows by modelling both within-view and between-view relationships among kernel values. We show, both on simulated data and real world data, that the proposed method outperforms existing techniques in the restricted settings where they are available, and extends applicability to new settings.
\end{abstract} 

\section{Introduction}

In recent years, many methods have been proposed for multi-view learning, i.e,  learning with data collected from multiple sources or \enquote{views} to utilize  the complementary information in them. Kernelized methods capture the similarities among data points in a kernel matrix. The multiple kernel learning (MKL) framework  \citep[c.f.][]{gon11} is a popular way to accumulate information from multiple data sources, where kernel matrices built on features from individual views are combined for better learning.  Commonly in MKL methods, it is assumed that  full kernel matrices for each view are available. However, in partial data analytics, it is common that information from some sources are not available for some data points.  

The incomplete data problem exists in a wide range of fields, including social sciences, computer vision, biological systems, and remote sensing. For example, in remote sensing, some sensors can go off for periods of time, leaving gaps to data. A second example is that when integrating legacy data sets, some views may not available for some data points, because  integration needs  were not considered when originally collecting and storing the data. For instance, gene expression may have been measured for some of the biological samples, but not for others, and as biological sample material has been exhausted, the missing measurements cannot be made any more. On the other hand, some measurements may be too expensive to repeat for all samples; for example, patient's genotype may be measured only if a particular condition holds. All these examples introduce missing views, i.e, all features of a view for a data point can be  missing simultaneously. 

{\bf Novelty in problem definition:} Previous methods for kernel completion have addressed single view kernel completion assuming individual missing values \citep{gra02,pai10}, or required at least one complete kernel with a full eigensystem to be used as an auxiliary data source \cite{tsu03}, or assumed a linear kernel approximation \cite{lia15}. Due to absence  of full rows/columns in the incomplete kernel matrices, no existing single-view kernel completion method \citep{gra02,pai10} can be applied to complete  kernel matrices of individual views independently. In the multi-view setting, \citet{tsu03} have proposed an expectation maximization based method to complete an incomplete kernel matrix for a view, with the help of a complete kernel matrix from another view. As it requires a full eigensystem of the auxiliary full kernel matrix, that method cannot be used to complete a kernel matrix with missing rows/columns when no other auxiliary complete kernel matrix is available.  On the other hand, \citet{lia15} proposed a generative model based method which approximates the similarity matrix for each view as a linear kernel in some low dimensional space. Therefore, it is unable to model highly non-linear kernels such as RBFs. 


 
{\bf Contribution:} In this paper, we propose a novel method to complete all incomplete kernel matrices collaboratively, by learning both between-view and within-view relationships among the kernel values (Figure \ref{fig:model}). We model between-view relationships in the following two ways: (1) Initially, adapting the  strategies from multiple kernel learning \cite{arg05,cor12} we complete kernel matrices, by expressing individual normalized kernel matrices corresponding to each view as a convex combination of normalized kernel matrices of other views. (2) Second, to model relationships between kernels having different eigensystems we propose a novel approach of restricting the local embedding of one view in to the convex hull of local embeddings of other views.

For within-view learning, we begin from the concept of local linear embedding \citep{row00}, 
reconstructing each feature representation for a kernel as a sparse linear combination of other available feature representations or \enquote{basis} vectors in the same view. We assume the local embeddings, i.e., the reconstruction weights and the basis vectors for reconstructing each samples, are similar across views. In this approach, the non-linearity of kernel functions of individual views is also preserved in the basis vectors.
The idea of restricting a kernel matrix into the convex hull of other kernel matrices has  already been used in the field of multiple kernel learning \citep{arg05} and kernel target alignment \citep{cor12}, where all kernel matrices have so far been assumed complete. In this paper we apply this idea for completing kernel matrices of individual views with the help of all incomplete kernel matrices of other views, by simultaneously completing all of them. The concept of local linear embeddings has previously been used in another application, namely multi-view clustering \citep{she13}, however with the restriction that all views have exactly the same embeddings or reconstruction weights. Our method is able to  model not only exactly the same but also similar embeddings of different views, which leads to improved performance. 




\section{Multi-view kernel completion}
\begin{figure*}[ht]
    \centering
    \includegraphics[width=0.8\textwidth]{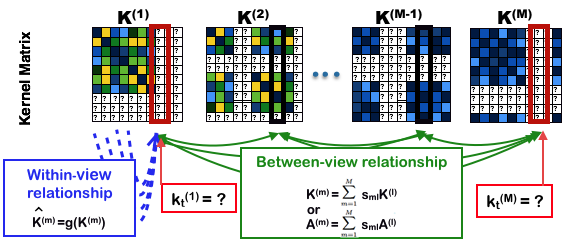}
    \caption{We assume $N$ data samples with $M$ views, with a few samples missing from each individual view, and consequently corresponding rows and columns are missing (denoted by '?') in kernel matrices ($\bK^{(m)}$). The proposed method predicts the missing kernel rows/columns (e.g., the $t^{th}$ column in views 1 and m  with the help of other samples of the same view (within-view relationship, blue arrows) and the corresponding sample in other views (between-view relationship, green arrows).  }
    \label{fig:model}
\end{figure*}



We assume $N$ data samples $\bX =\{ \bx_1,\ldots,\bx_N\}$ from a multi-view input space $\XX = \XX^{1} \times \dots \times \XX^{(M)}$, where $\XX^{(m)}$ is the input space generating the $m^{th}$ view. We denote by $\bX^{(m)}=\{ \bx^{(m)}_1,\ldots,\bx^{(m)}_N\}$, $\forall$  $m = 1,\ldots , M$, the set of samples for the $m^{th}$ view, where $\bx^{(m)}_i \in \XX^{(m)}$ is the $i^{th}$ observation in the $m^{th}$ view and $\XX^{(m)}$ is the input space. Considering an implicit mapping of the observations of the $m^{th}$ view to an inner product space $\FF^{(m)}$ via a mapping  $\phi^{(m)} : \XX^{(m)} \rightarrow \FF^{(m)}$,  and following  the usual recipe for kernel methods \citep{bac04}, we specify the kernel as the inner product in $\FF^{(m)}$. The kernel value between the $i^{th}$ and $j^{th}$ data points is defined as ${k}^{(m)}_{ij}=  \langle \phi_i^{(m)},\phi_j^{(m)} \rangle $, where  $\phi_i^{(m)} = \phi^{(m)}(\bx_i^{(m)})$ and $k_{ij}^{(m)}$ is an element of $\bK^{(m)}$, the kernel Gram matrix for the set $\bX^{(m)}$. 

In this paper we assume that all samples are not available in all views. Let $\bI = [1,\ldots, N] $ be the set of  indices of all data points and  $\bI^{(m)}$ be the set of indices of all available data points in the $m^{th}$ view. Hence for each view, only a kernel sub-matrix ($\bK^{(m)}_{\bI^{(m)}\bI^{(m)}}$) corresponding to the rows and columns indexed by $\bI^{(m)}$ is known. Our aim is to predict a complete positive semi-definite kernel matrix ($\hat{\bK}^{(m)} \in \R^{N \times N}$)  corresponding to  each view, where the sub-matrix $\hat{\bK}^{(m)}_{\bI^{(m)}\bI^{(m)}}$ of it  is known to be equal to $\bK^{(m)}_{\bI^{(m)}\bI^{(m)}}$. This calls for predicting missing ($t^{th}$) rows/column of $\hat{\bK}^{(m)}$, for all $t \in \{ \bI/\bI^{(m)}\}$.

Our approach for predicting $\hat{\bK}^{(m)}$ is based on  exploiting information available within the same view but in particular in other views. The proposed method predicts missing values by learning both between-view and within-view relationships among the kernel values    (Figure \ref{fig:model}).

\subsection{Within-view kernel relationships}

 
 For within-view learning, relying on the concept of local linear embedding \citep {row00}, we reconstruct the feature map of $t^{th}$ data point $\phi_t^{(m)}$ by a sparse linear combination of known data samples $$\hat{\phi}^{(m)}_t= \sum_{i \in \bI^{(m)}} a_{it}^{(m)} \phi_i^{(m)}$$ where $ a_{it}^{(m)} \in \R$ is the reconstruction weight of the $i^{th}$ feature representation for representing the $t^{th}$ sample. Hence, approximated kernel values can be expressed as 
\begin{equation}
 \hat{k}_{tt'} = \langle \hat{\phi}^{(m)}_t,\hat{\phi}^{(m)}_{t'} \rangle = \sum_{i,j \in \bI^{(m)}} a_{it}^{(m)} a_{jt'}^{(m)} \langle \phi_i^{(m)} , \phi_j^{(m)} \rangle. \nonumber 
 \label{eqn:ktt}
 \end{equation}
 We collect all reconstruction weights of a view into the matrix $\bA^{(m)} = \left(a_{ij}\right)_{i,j=1}^N$. Further, by $\bA^{(m)}_{I^{(m)}}$ we denote the submatrix of  $\bA^{(m)}$ containing the rows indexed by $I^{(m)}$, the known data samples in the view. Thus the reconstructed kernel matrix $\hat{\bK}^{(m)}$ can be written as
\begin{eqnarray}
    \hat{\bK}^{(m)} =  \bA^{(m)^T}_{I^{(m)}} \bK^{(m)}_{I^{(m)}I^{(m)}} \bA^{(m)}_{I^{(m)}} = g(\bK^{(m)})\,\,\,\,
    \label{eqn:app}
\end{eqnarray}
Note that $\hat{\bK}^{(m)}$ is positive semi-definite when $\bK^{(m)}$ is positive semi-definite.  Therefore, with the help of this approximation one can avoid introducing an explicit positive semi-definiteness constraint in the optimization.

Intuitively, the reconstruction weights are used to extend the known part of the kernel to the unknown part, in other words, the unknown part is assumed to reside within the span of the known part. 

We further assume that in each view there exists a sparse embedding in $\FF^{(m)}$, given by a small set of samples $\bB^{(m)} \subset \bI^{(m)}$, called a basis set, that is able to represent all possible feature representations in that particular view. Thus the non-zero reconstruction weights are confined to the basis set: $a_{ij}^{(m)} \neq 0$ only if $i \in \bB^{(m)}$. To select such a sparse set of reconstruction weights, we regularize the reconstruction weights by the $l_{2,1}$ norm \citep{arg06} of the reconstruction weight matrix, 
$$\|\bA^{(m)}_{I^{(m)}}\|_{2,1} = \sum_{i \in \bI^{(m)}}\sqrt{\sum_{j \in \bI} (a_{ij}^{(m)})^2}.$$
Finally, for the known part of the kernel, we add the additional objective that the reconstructed kernel values closely approximate the observed values. 
For this end, we define a loss function measuring the within-view approximation error for the $m^{th}$ view as 
       \begin{eqnarray}
       Loss_{within}^{(m)} = \|\hat{\bK}^{(m)}_{\bI^{(m)}\bI^{(m)}}- \bK^{(m)}_{\bI^{(m)}\bI^{(m)}}\|^2_2.
         \label{eqn:loss_pred}
     \end{eqnarray}  
 We note that without the $\ell_{2,1}$ regularization, the above approximation loss would be trivially optimized by choosing $\bA^{(m)}_{\bI^{(m)}\bI^{(m)}}$ as the identity matrix. The  $\ell_{2,1}$ regularization will have the effect of zeroing out some of the diagonal values and introducing non-zeros to the submatrix $\bA^{(m)}_{\bI^{(m)}\bI/\bI^{(m)}}$, corresponding to the rows and columns indexed by $\bI^{(m)}$ and $\bI/\bI^{(m)}$ respectively.

We note that the above approach differs from the the sparse kernel approximation methods, such random Fourier features \citep{rah07} and the Nystr{\" o}m method \citep{ dri05} which have been successfully applied to efficient kernel learning.
Namely, these methods find global basis vectors spanning the kernel whereas our method finds local reconstruction weights for the data samples. In addition, we aim to optimize these reconstruction weights using the other views for optimal kernel approximation.


\subsection{Between-view kernel relationships}

 For a completely missing row or column of a kernel matrix, there is not enough information available for completing it within the same view, and hence the completion needs to be based on other information sources, in our case the other views where the corresponding kernel parts are known.  In the following, we introduce two approaches for relaying information of the other view for completing the unknown rows/columns of a particular view. The first technique is based on learning a convex combination of the kernels, extending the multiple kernel learning \citep{arg05,cor12} techniques to kernel completion.  The second technique is based on learning reconstruction weights so that they share information between the views. 
 
{\bf Between-view learning of kernel values.}
 To learn between-view relationships we express the individual normalized kernel matrix corresponding to each view as a convex combination of normalized kernel matrices of the other views. Hence the proposed model learns kernel weights $\bS = (s_{ml})_{m,l=1}^M$ between all pairs of kernels $(m,l)$ such that
\begin{eqnarray}
\hat{\bK}^{(m)} &\approx& \sum_{l=1,l \ne m}^{M} s_{ml} \hat{\bK}^{(l)},
\label{eqn:convhul-k}
\end{eqnarray}  
where the kernel weights are confined to a convex combination 
$\mS =\{\bS | s_{ml} \ge 0, \,\, \sum_{l=1,l \ne m}^{M} s_{ml} =1\}$. The kernel weights then can flexibly pick up a subset of relevant views to the current view $m$. Previously, \citet{arg05} have proposed a method for learning kernels by restricting the search in the convex hull of a set of given kernels to learn  parameters of individual kernel matrices. Here, we apply the idea to kernel completion, which has not been previously considered. We further note that kernel approximation as a convex combination has the interpretation of avoiding extrapolation in the space of kernels, and can be interpreted as a type of regularization to constrain the otherwise flexible set of PSD kernel matrices. 


To learn  the between-view relationships among kernels we define a between-view  loss for each of the $M$ views based on the above approximation as  
 \begin{eqnarray}
         Loss_{between}^{(m)}(\hat{\bK},\bS) = \|\hat{\bK}^{(m)}- \sum_{l=1,l \ne m}^{M} s_{ml} \hat{\bK}^{(l)}\|^2_2.
         \label{eqn:loss_inter}
     \end{eqnarray}  
{\bf Between-view learning of reconstruction weights.} 
In practical applications, the kernels arising in a multi-view setup might be very heterogeneous in their distributions. In such cases, it might not be realistic to find a convex combination of other kernels that are closely similar to the kernel of a given view. In particular, when the eigen-spectra of the kernels are very different, we expect achieving a low between-view loss (Equation (\ref{eqn:loss_inter})) to be hard to achieve.


For such cases, we propose an alternative approach, where instead of the kernel values, we assume that the basis sets and the reconstruction weights have between-view dependencies that we can learn. To capture the relationship, we assume the reconstruction weights in a view can be approximated by a convex combination of the reconstruction weights of the other views
\begin{eqnarray}
\bA^{(m)}_{\bI^{(m)}} \approx \sum_{l=1,l\ne m}^M s_{ml}\bA^{(l)}_{\bI^{(m)}},
\label{eqn:convhul-A}
\end{eqnarray}  
where $s_{ml}$ is defined in Equation (\ref{eqn:convhul-k}). 
This gives us between-view loss for reconstruction weights as
 \begin{eqnarray}
         Loss_{between}^{(m)}(\bA,\bS) =\|{\bA}^{(m)}_{\bI^{(m)}} -\sum_{l=1,l\ne m}^M s_{ml}\bA^{(l)}_{\bI^{(m)}} \|^2_2.
        \label{eqn:loss_interA}
\end{eqnarray}  
Given the above, the reconstructed kernel is thus given by
\begin{small}
\begin{align*}
   \hat{\bK}^{(m)} & =  \nonumber \\ 
    & \left(  \sum_{l=1,l\ne m}^M s_{ml}\bA_{\bI^{(m)}} ^{(l)^T}\right) \bK^{(m)}_{\bI^{(m)}\bI^{(m)}} \left(\sum_{l=1,l\ne m}^M s_{ml}\bA^{(l)}_{\bI^{(m)}} \right). \,\,\,\,
    \label{eqn:kernel_between}
\end{align*}
\end{small}

Learning the reconstruction weight matrix in multi-view setup has recently been considered by \citet{she13}. However, they assume the same global reconstruction weights in all views ($\bA = \bA^{(m)}$ for all $m$), which is in our view a very strong assumption. Thus, in our approach we allow the views to have different reconstruction weights, but assume a parameterized relationship learned from data.  

\section{Optimization problems}

Here we present the optimization problems for \textbf{Multi-view Kernel Completion} (\mkc),\ arising from the within-view and between-view kernel approximations described above.



{\bf \mkc\ using semi-definite programming (\psd):\ }
This is the most general case where we do not put any other restrictions on kernels of individual views, other than restricting them to be positive semi-definite kernels.
In this general case we propagate information from other views by learning between-view relationships depending on kernel values in Equation (\ref{eqn:convhul-k}). Hence, using Equations (\ref{eqn:loss_pred} and \ref{eqn:loss_inter}) we get
\begin{eqnarray}
\min_{\substack{\bS,\hat{\bK}^{(m)},\\ m=1,\ldots,M}} & & \sum_{m=1}^M \left(Loss_{within}^{(m)} +c Loss_{between}^{(m)}(\hat{\bK},\bS)\right) \nonumber \\
s.t. &&\hat{\bK}^{(m)} \succeq 0 \, \forall \, m=1,\ldots,M \nonumber \\
&& \bS \in \mS.
\label{eqn:mkc_sdp}
\end{eqnarray}
We solve this non-convex optimization problem by iteratively solving it for $\bS$ and $\hat{\bK}^{(m)}$ using block-coordinate descent. For a fixed $\bS$, to update the $\hat{\bK}^{(m)}$'s  we need to solve a semi-definite program with $M$ positive constraints. 

{\bf \mkc\ using heterogeneous embeddings (\emht):\ }
An optimization problem with $M$ positive semi-definite constraints is inefficient for even a data set of size $100$. To avoid solving the SDP in each iteration we assume a kernel approximation (Equation \ref{eqn:app}). When kernel functions in different views are not the same and kernel matrices in different views have different eigen-spectra, it is good to learn relationships among underlying embeddings of different views (Equation \ref{eqn:convhul-A}), instead of the actual kernel values. Hence, using  Equations (\ref{eqn:loss_pred}, \ref{eqn:app} and \ref{eqn:loss_interA}) along with $l_{2,1}$ regularization on $\bA^{(m)}$, we get
\begin{eqnarray}
\min_{ \substack{ \bS, \bA^{(m)},\hat{\bK}^{(m)}, \\ \forall m=1,\ldots,M}} && \hspace{-0.5cm} \sum_{m=1}^M \left(Loss_{within}^{(m)} +c_1 Loss_{between}^{(m)}(\bA,\bS) \right) \nonumber \\
&&  +c_2\sum_{m=1}^M \|\bA^{(m)}\|_{2,1}\nonumber \\
s.t
&&\hspace{-0.5cm}\hat{\bK}^{(m)} =  \bA^{(m)^T}_{I^{(m)}} \bK^{(m)}_{I^{(m)}I^{(m)}} \bA^{(m)}_{I^{(m)}} \nonumber \\
&& \hspace{-0.5cm}\bS \in \mS 
\label{eqn:mkc_emht}
\end{eqnarray}
{\bf \mkc\ using kernel approximation (\mkckernel):\ }
To study the advantages of learning relationships using underlying embeddings instead of using kernel values, we consider the following optimization problem. In this case the kernel is approximate but between-view relationships are learnt on kernel values using Equation (\ref{eqn:loss_inter}):
\begin{eqnarray}
\min_{\substack{\bS,\bA^{(m)},\hat{\bK}^{(m)},\\ \forall m=1,\ldots,M}} & & \hspace{-0.5cm} \sum_{m=1}^M \left(Loss_{within}^{(m)} +c_1 Loss_{between}^{(m)}(\hat{\bK},\bS)\right) \nonumber \\
&&  +c_2\sum_{m=1}^M \|\bA^{(m)}\|_{2,1}\nonumber \\
s.t&& \hspace{-0.5cm} \hat{\bK}^{(m)} =  \bA^{(m)^T}_{I^{(m)}} \bK^{(m)}_{I^{(m)}I^{(m)}} \bA^{(m)}_{I^{(m)}}  \nonumber \\
&& \hspace{-0.5cm} \bS \in \mS
\label{eqn:mkc_app}
\end{eqnarray}
We solve all the above-mentioned  non-convex optimization problems with $l_{2,1}$ regularization by sequentially updating $\bS$ and $\bA^{(m)}$. In each iteration $\bS$ is updated by solving a quadratic program and for each  $m$, $\bA^{(m)}$ is updated using proximal gradient descent.
%


\section{Algorithm}
\label{sec:algo}
Algorithm \ref{algo:emht} describes the main algorithm to solve \emht\ (Equation \ref{eqn:mkc_emht}). Algorithms for solving the other variants are similar and are presented in the supplementary material for lack of space. 

\begin{algorithm}
\caption{\hspace{-0.1cm}.  \emht \begin{small}$\left(\bK^{(m)},\bI^{(m)}, \forall m \in[1,\ldots,M] \right)$ \end{small}}\label{algo:emht}
\begin{algorithmic}
\STATE Initialization: 
\STATE $s^0_{mm}=0$, $s^0_{ml}=\frac{1}{M-1}$, 
\STATE  $\bA^{(m)^0}_{\bI^{(m)}\bI^{(m)}}$ = Identity matrix and
\STATE $a^{(m)^0}_{tt'} \sim uniform(-1,1)$ for all $t,t' \notin \bI^{(m)}$ 
\vspace{0.5cm}
\REPEAT 
\vspace{0.1cm}

\FOR {m=1 to M } 
\FOR {t=1 to N}
\STATE \hspace{-0.1cm}$ \ba_t^{(m)^k} = max\left(0, (1-\frac{\lambda c_2}{\|\Delta \ba_t^{(m)^{k-1}}\|_2}) \Delta \ba_t^{(m)^{k-1}}\right)$
\vspace{0.2cm}
    \STATE \hspace{2cm} [according to Equations (\ref{eqn:prox}, \ref{eqn:prox_m})]
\STATE \hspace{3cm} [$\lambda$ is fixed by line search]
\ENDFOR
\ENDFOR
\vspace{0.1cm}

\FOR {m=1  to M}
\STATE 
$\bs_m^k = \arg\min_{\bs_m}  Sobj_{\bA^{(m),m=1,\ldots,M}}^k(\bs_m) \nonumber $
\STATE \hspace{3cm}[according to Equation (\ref{eqn:mkc_emht_S})]
\ENDFOR
\vspace{0.1cm}

\UNTIL{converge}
\end{algorithmic}
\end{algorithm}

Substituting $\hat{\bK}^{(m)} =  \bA^{(m)^T}_{\bI^{(m)}} \bK^{(m)}_{\bI^{(m)}\bI^{(m)}} \bA^{(m)}_{\bI^{(m)}}$, the Equation (\ref{eqn:mkc_emht}) has two sets of unknowns, $\bS$ and the $\bA^{(m)}$'s. We update  $\bA^{(m)}$ and $\bS$ in an iterative manner. In the $k^{th}$ iteration for a fixed $\bS^{k-1}$ from the previous iteration, to update $\bA^{(m)}$'s  we need to solve following for each  $m$:
\begin{eqnarray}
 \bA^{(m)^k} = \arg\min_{\bA^{(m)}}   Aobj_{\bS}^k(\bA^{(m)}) +c_2 \Omega(\bA^{(m)}) \nonumber
\end{eqnarray}
\label{eqn:mkc_emht2}
where \begin{small} $ \Omega(\bA^{(m)})=\|\bA^{(m)}\|_{2,1}$ \end{small} and \begin{small} $Aobj_{\bS}^k(\bA^{(m)})=\|\bK^{(m)}_{\bI^{(m)}\bI^{(m)}}- \left[\bA^{(m)^T}_{\bI^{(m)}}\bK^{(m)}_{\bI^{(m)}\bI^{(m)}}\bA^{(m)}_{\bI^{(m)}}\right]_{\bI^{(m)}\bI^{(m)}}\|_2^2 +c_1 \sum_{m=1}^M \|\bA^{(m)} - \sum_{l=1, l\ne m}^{M}s^{k-1}_{ml}\bA^{(l)}  \|_2^2$ \end{small}.

Instead of solving this problem in each iteration we update $\bA^{(m)}$ using proximal gradient descent. Hence, in each iteration,
\begin{eqnarray}
    \bA^{(m)^k} \hspace{-0.1cm}= Prox_{\lambda c_2\Omega}\left(\bA^{(m)^{k-1}}-\lambda \partial Aobj_{\bS}^k(\bA^{(m)^{k-1}}) \right)&&
  \label{eqn:prox}
\end{eqnarray}
where $\partial Aobj_{\bS}^k(\bA^{(m)})$ is the differential of $Aobj_{\bS}^{k}(\bA^{(m)})$ at $\bA^{(m)^{k-1}}$ and $\lambda$ is the step size which is decided by a line search. 
In Equation (\ref{eqn:prox}) each row of $\bA^{(m)}$ (i.e., $\ba_{t}^{(m)}$) can be solved independently and we apply a proximal operator on each row. Following \citet{bach11c}, the  solution of Equation (\ref{eqn:prox}) is 
\begin{equation}
    \ba_t^{(m)^k} = max\left(0, (1-\frac{\lambda c_2}{\|\Delta \ba
    _t^{(m)^{k-1}}\|_2}) \Delta \ba_t^{(m)^{k-1}}\right),
\label{eqn:prox_m}
\end{equation}
where $\Delta \ba_t^{(m)^{k-1}}$ is  the $t^{th}$ row of \begin{small} $\left(\bA^{(m)^{k-1}}-\lambda \partial Aobj_{\bS}^k(\bA^{(m)^{k-1}})\right)$ \end{small}.

Again, in the $k^{th}$ iteration, for fixed $\bA^{(m)^k}$'s,  the $\bS$ is updated by  independently updating each row ($\bs_m$) through solving the  following Quadratic Program:
\begin{eqnarray}
\bs_m^{k} &=& \arg\min_{\bs_m}  Sobj_{\bA^{(m),m=1,\ldots,M}}^k(\bs_m) \nonumber \\
s.t&& \sum_{l\ne m} s_{ml} =1, \nonumber \\
&& s_{ml} \ge 0 \,\, \forall l
\label{eqn:mkc_emht_S}
\end{eqnarray}
where 
\begin{small}
$$Sobj_{\bA^{(m),m=1,\ldots,M}}^k(\bs_m)=\|\bA^{(m)^k}-\sum_{l=1, l\ne m}^{M}s_{ml}\bA^{(l)^k}\|_2^2.$$
\end{small} 
 
\section{Experiments}

 We  apply the proposed \mkc\ method on a variety of data sets, with different types of kernel functions in different views, along with different amounts of missing data points. The objectives of our experiments  are: (1) to compare the performance of \mkc\ against other existing methods in terms of the ability  to predict the missing kernel rows, (2) to empirically show that the proposed kernel approximation with the help of the reconstruction weights also improves running-time over the \psd\ method. 


\subsection{Experimental setup}
\subsubsection{Data sets:} To  evaluate  the performance  of  our  method,  we  used 4 simulated data sets with 100 data points and 5 views, as well as two real-world multi-view data sets: (1) Dream Challenge 7 data set (DREAM) \citep{dae13, hei12} and (2) Reuters RCV1/RCV2 multilingual data \citep{ami09}. 

\begin{figure}[ht]
       \centering
       \includegraphics[width= 0.5 \textwidth]{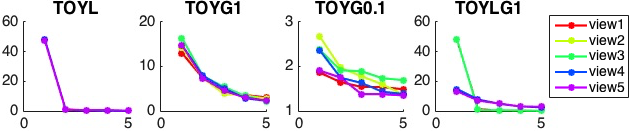} \\
       \includegraphics[width = 0.5 \textwidth]{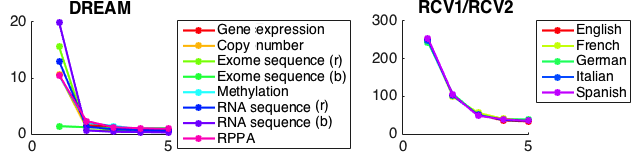}
       \caption{Eigen-spectra of kernel matrices of the different views in the data sets. 
       Each coloured line in a plot shows the eigen-spectrum in one view. Here, (r) indicates use of Gaussian kernel on real values whereas (b) indicates use of Jaccard's kernel on binarized values.}
       \label{fig:eigspec}
   \end{figure}

{\bf Synthetic data sets:}We followed the following steps to simulate our synthetic data sets:
\begin{itemize}
    \item[1] We generated the first 10 points ($\bX_{\bB^{(m)}}^{(m)}$) for each view, where $\bX_{\bB^{(m)}}^{(1)}$ and $\bX_{\bB^{(m)}}^{(2)}$ are uniformly distributed in $[-1,1]^{5}$ and $\bX_{\bB^{(m)}}^{(3)}$, $\bX_{\bB^{(m)}}^{(4)}$ and, $\bX_{\bB^{(m)}}^{(5)}$ are uniformly distributed in $[-1,1]^{10}$.
    \item[2] These 10 data points were used as basis sets for each view, and further 90 data points in each view were generated by $\bX^{(m)}=\bA^{(m)} \bX_{\bB^{(m)}}^{(m)}$, where the $\bA^{(m)}$ are uniformly distributed random matrices $ \in \R^{90 \times 10}$. We chose $\bA^{(1)}=\bA^{(2)}$ and $\bA^{(3)}=\bA^{(4)} = \bA^{(5)}$. 
    \item[3] Finally, $\bK^{(m)}$ was generated from $\bX^{(m)}$ by using different kernel functions for different data sets as follows: 
    \begin{itemize}
        \item TOYL : Linear kernel for all views 
        \item TOYG1 and TOYG0.1 : Gaussian kernel for all views where the kernel with of the Gaussian kernel are $1$ and $0.1$ respectively.
        \item TOYLG1 : Linear kernel for the first 3 views and Gaussian kernel for the last two views with the kernel width $1$. Note that with this selection view 3 shares reconstruction weights with view 4 and 5, but has the same kernel as views 1 and 2.
    \end{itemize}
   Figure \ref{fig:eigspec} shows the eigen-spectra of kernel matrices of the 5 views for all simulated data sets.
   \end{itemize}
   
   {\bf The Dream Challenge 7 data set (DREAM):} For Dream Challenge 7, genomic characterizations of multiple types on 53 breast cancer cell lines are provided. They consist of DNA copy number variation, transcript expression values, whole exome sequencing, RNA sequencing data, DNA methylation data and RPPA protein quantification measurements. In addition, some of the views are missing for some cell lines. For 25 data points all 6 views are available. For all the 6 views,  we calculated Gaussian kernels after normalizing the data sets. We generated other two kernels by using Jaccard's kernel function over binarized exome data and RNA sequencing data. Hence, the final data set has 8 kernel matrices. Figure \ref{fig:eigspec} shows the eigen-specta of the kernel matrices of all views.
    
    {\bf RCV1/RCV2:}  Reuters RCV1/RCV2 multilingual data set contains aligned documents for 5 languages (English, French, Germany, Italian and Spanish). Originally the documents are in any one of these languages and then corresponding documents for other views have been generated by machine translations of the original document. For our experiment, we randomly selected 1500 documents which were originally in English. The latent semantic kernel \citep{cri02} is used for all languages.
     
   \begin{table*}[ht]
       \centering
       \scriptsize
       \begin{tabular}{|l|c|c|c|c|c|c|}
      \multicolumn{7}{c}{Number of missing views = 1 (TOY and RCV1/RCV2) and 1 (DREAM)}\\ \hline
      Algorithm & TOYL &TOYG1 &TOYG0.1&TOYLG1&DREAM&RCV1/RCV2 \\ \hline  
 \emht\ \hspace{-0.2cm}& {\bf  0.07 ($\pm$  0.09)} 	\hspace{-0.2cm}&   7.40 ($\pm$  9.20) 	\hspace{-0.2cm}&  84.91 ($\pm$  5.18) 	\hspace{-0.2cm}&   4.50 ($\pm$  6.72) 	\hspace{-0.2cm}&  {\bf 13.36 ($\pm$ 26.53)} &   1.79( $\pm$    0.89) 	\\  \mkckernel\ \hspace{-0.2cm}&   0.09 ($\pm$  0.10) 	\hspace{-0.2cm}&  {\bf 5.02 ($\pm$  3.60) }	\hspace{-0.2cm}&  76.24 ($\pm$ 10.59) 	\hspace{-0.2cm}&   {\bf 2.11 ($\pm$  3.40) }	\hspace{-0.2cm}&  14.46 ($\pm$ 28.39) \hspace{-0.2cm}&   {\bf 1.15( $\pm$    0.48) }	\\  
 \psd\ \hspace{-0.2cm}&   0.22 ($\pm$  0.32) 	\hspace{-0.2cm}&  11.29 ($\pm$  6.29) 	\hspace{-0.2cm}&   {\bf 7.83 ($\pm$  5.46) }	\hspace{-0.2cm}&   6.06 ($\pm$  7.84) 	\hspace{-0.2cm}&  20.19 ($\pm$ 41.28) \hspace{-0.2cm}&  - 	\\  
 \emhm\ \hspace{-0.2cm}&   0.19 ($\pm$  0.18) 	\hspace{-0.2cm}&  27.54 ($\pm$ 14.38) 	\hspace{-0.2cm}&  86.08 ($\pm$  6.34) 	\hspace{-0.2cm}&   8.93 ($\pm$ 11.86) 	\hspace{-0.2cm}&  16.12 ($\pm$ 30.27) \hspace{-0.2cm}&	 3.27( $\pm$    1.26)\\  
  \hline
  \emb\ \hspace{-0.2cm}&  20.65 ($\pm$ 41.08) 	\hspace{-0.2cm}& 554.08 ($\pm$  90.00) 	\hspace{-0.2cm}&  31.23 ($\pm$ 37.02) 	\hspace{-0.2cm}& 759.74 ($\pm$  90.00) 	\hspace{-0.2cm}&  14.78 ($\pm$ 32.93) \hspace{-0.2cm}& 23.38( $\pm$  29.00) 	\\  
  \knn\ \hspace{-0.2cm}&   0.34( $\pm$    0.53) 	\hspace{-0.2cm}&  42.89( $\pm$   27.93) 	\hspace{-0.2cm}&  62.69( $\pm$    8.77) 	\hspace{-0.2cm}&  11.27( $\pm$   15.53) 	\hspace{-0.2cm}&  14.94( $\pm$   25.29) 	\hspace{-0.2cm}&   5.79( $\pm$    2.65) 	\\  
 \wknn\ \hspace{-0.2cm}&   0.34( $\pm$    0.53) 	\hspace{-0.2cm}&  45.47( $\pm$   29.50) 	\hspace{-0.2cm}&  62.80( $\pm$    8.86) 	\hspace{-0.2cm}&  15.30( $\pm$   20.15) 	\hspace{-0.2cm}&  15.00( $\pm$   25.35) 	\hspace{-0.2cm}&   5.91( $\pm$    2.71) 	\\  
\hline
\multicolumn{7}{c}{Number of missing views = 2 (TOY and RCV1/RCV2) and 3 (DREAM)}\\ \hline
Algorithm & TOYL &TOYG1 &TOYG0.1&TOYLG1&DREAM&RCV1/RCV2 \\ \hline
\emht\ \hspace{-0.2cm}&   0.08 ($\pm$  0.07) 	\hspace{-0.2cm}&   9.43 ($\pm$  6.72) 	\hspace{-0.2cm}&  86.72 ($\pm$  3.34) 	\hspace{-0.2cm}&   {\bf 3.26 ($\pm$  5.07)} 	\hspace{-0.2cm}& {\bf 16.13 ($\pm$ 28.29)} \hspace{-0.2cm}&   2.74( $\pm$    0.85) 	\\  
 \mkckernel\ \hspace{-0.2cm}& { \bf  0.07 ($\pm$  0.05) }	\hspace{-0.2cm}&  { \bf 6.89 ($\pm$  3.44) }	\hspace{-0.2cm}&  84.40 ($\pm$  9.04) 	\hspace{-0.2cm}&   4.01 ($\pm$  6.03) 	\hspace{-0.2cm}&  17.51 ($\pm$ 27.65)  \hspace{-0.2cm}& {\bf 1.61( $\pm$    0.65) }	\\  
 \psd\ \hspace{-0.2cm}&   0.34 ($\pm$  0.39) 	\hspace{-0.2cm}&  19.87 ($\pm$ 13.88) 	\hspace{-0.2cm}&  { \bf 18.30 ($\pm$ 12.94) }	\hspace{-0.2cm}&  37.88 ($\pm$ 49.58) 	\hspace{-0.2cm}&  32.86 ($\pm$ 51.73) \hspace{-0.2cm}&   - 	\\  
 \emhm\ \hspace{-0.2cm}&   0.14 ($\pm$  0.09) 	\hspace{-0.2cm}&  29.69 ($\pm$  9.85) 	\hspace{-0.2cm}&  96.19 ($\pm$  1.60) 	\hspace{-0.2cm}&  13.78 ($\pm$ 21.33) 	\hspace{-0.2cm}&  18.33 ($\pm$ 29.56) \hspace{-0.2cm}& 	 2.71( $\pm$    0.71)\\  
 \hline
 \emb\ \hspace{-0.2cm}&  28.66 ($\pm$ 42.28) 	\hspace{-0.2cm}& 202.58 ($\pm$  339.02) 	\hspace{-0.2cm}&  61.87 ($\pm$ 50.48) 	\hspace{-0.2cm}& 298.57 ($\pm$  281.79) 	\hspace{-0.2cm}&  25.98 ($\pm$ 63.70) 	\hspace{-0.2cm}&  27.83 ($\pm$ 13.70)    \\  
 \knn\ \hspace{-0.2cm}&   0.26( $\pm$    0.26) 	\hspace{-0.2cm}&  52.18( $\pm$   16.34) 	\hspace{-0.2cm}&  97.65( $\pm$   11.62) 	\hspace{-0.2cm}&  19.64( $\pm$   25.62) 	\hspace{-0.2cm}&  22.04( $\pm$   30.07) 	\hspace{-0.2cm}&   7.47( $\pm$    2.38) 	\\  
 \wknn\ \hspace{-0.2cm}&   0.26( $\pm$    0.26) 	\hspace{-0.2cm}&  54.94( $\pm$   16.02) 	\hspace{-0.2cm}&  98.24( $\pm$   11.17) 	\hspace{-0.2cm}&  20.90( $\pm$   27.16) 	\hspace{-0.2cm}&  22.20( $\pm$   30.26) 	\hspace{-0.2cm}&   7.61( $\pm$    2.40) 	\\ 
 \hline
\multicolumn{7}{c}{Number of missing views = 3 (TOY and RCV1/RCV2) and 5 (DREAM)}\\ \hline
Algorithm & TOYL &TOYG1 &TOYG0.1&TOYLG1&DREAM&RCV1/RCV2 \\ \hline
\emht\ \hspace{-0.2cm}&  {\bf 0.05 ($\pm$  0.04)} 	\hspace{-0.2cm}&  12.87 ($\pm$  3.40) 	\hspace{-0.2cm}&  89.88 ($\pm$  3.26) 	\hspace{-0.2cm}&   {\bf 5.13 ($\pm$  7.17) }	\hspace{-0.2cm}&  {\bf 20.04 ($\pm$ 30.58) } 	\hspace{-0.2cm}& {\bf  1.69( $\pm$    0.85)}	\\  
\mkckernel\ \hspace{-0.2cm}&   0.10 ($\pm$  0.05) 	\hspace{-0.2cm}& {\bf 12.04 ($\pm$  3.71) }	\hspace{-0.2cm}&  89.69 ($\pm$  5.54) 	\hspace{-0.2cm}&   5.72 ($\pm$  7.88) 	\hspace{-0.2cm}&  20.43 ($\pm$ 30.39) \hspace{-0.2cm}&   2.91( $\pm$    3.15) 	\\  
 \psd\ \hspace{-0.2cm}&   0.41 ($\pm$  0.35) 	\hspace{-0.2cm}&  86.21 ($\pm$ 55.84) 	\hspace{-0.2cm}& {\bf 17.59 ($\pm$  9.37)} 	\hspace{-0.2cm}& 438.92 ($\pm$  624.21) 	\hspace{-0.2cm}&  97.79 ($\pm$ 89.51) \hspace{-0.2cm}&  -	\\  
  \emhm\ \hspace{-0.2cm}&   0.16 ($\pm$  0.10) 	\hspace{-0.2cm}&  32.70 ($\pm$ 10.63) 	\hspace{-0.2cm}&  95.43 ($\pm$  1.75) 	\hspace{-0.2cm}&  15.91 ($\pm$ 23.31) 	\hspace{-0.2cm}&  22.13 ($\pm$ 33.29) \hspace{-0.2cm}&  2.45( $\pm$    1.54)  	\\  
 \hline
 \emb\ \hspace{-0.2cm}&  21.46 ($\pm$ 40.73) 	\hspace{-0.2cm}& 101.87 ($\pm$ 63.31) 	\hspace{-0.2cm}& 554.08 ($\pm$  90.00) 	\hspace{-0.2cm}& 231.98 ($\pm$  416.30) 	\hspace{-0.2cm}&  60.47 ($\pm$  245.88) \hspace{-0.2cm}&  29.76 ($\pm$ 12.28)	\\  
\knn\ \hspace{-0.2cm}&   0.39( $\pm$    0.33) 	\hspace{-0.2cm}&  62.32( $\pm$   14.94) 	\hspace{-0.2cm}& 112.79( $\pm$   13.22) 	\hspace{-0.2cm}&  24.93( $\pm$   33.57) 	\hspace{-0.2cm}&  27.54( $\pm$   36.88) 	\hspace{-0.2cm}&   8.66( $\pm$    1.99) 	\\  
 \wknn\ \hspace{-0.2cm}&   0.38( $\pm$    0.33) 	\hspace{-0.2cm}&  66.94( $\pm$   14.74) 	\hspace{-0.2cm}&  97.96( $\pm$    4.52) 	\hspace{-0.2cm}&  27.48( $\pm$   36.90) 	\hspace{-0.2cm}&  27.57( $\pm$   36.70) 	\hspace{-0.2cm}&   8.85( $\pm$    1.99) \\
 \hline 	
       \end{tabular}
       \caption{ Average relative error percentage (Equation (\ref{eqn:are})). The smallest ARE for each setup are boldfaced. The figures are ARE averaged over all views and 5 random validation and test partitions with different missing entries (standard deviation in parentheses).}
       \label{tab:rmse}
   \end{table*}
    
   \subsubsection{Evaluation setup}
   Each of the data sets was partitioned into tuning and test sets.  The missing views were introduced in these partitions independently. To induce missing views, we randomly selected data points from each partition, a few views for each of them, and deleted the corresponding rows and columns from the kernel matrices. The tuning set was used for parameter tuning. All the results have been reported on the test set which was independent of the tuning set. 
   
   For all 4 synthetic data sets as well as  RCV1/RCV2 we chose $40\%$ of the data samples as the tuning set, and the rest $60\%$ were used for testing. For the DREAM data set these partitions were $60\%$ for tuning and $40\%$ for testing.
   
   We generated versions of the data with different amounts of missing values. For the first test case, we deleted 1 view from each selected data point in each data set.  In the second test case, we removed 2 views for TOY and RCV1/RCV2 data sets and 3 views for DREAM. For the third one we deleted 3 views among 5 views per selected data point in TOY and RCV1/RCV2, and 5 views among 8 views per selected data point in DREAM. 
   
   We repeated all our experiments for 5 random tuning and test partitions with different missing entries and report the average performance on them.

   \subsubsection{Compared methods}
%

   We compared performance of the proposed methods, \emht,\ \mkckernel,\ \psd,\ with {\it $k$ nearest neighbour} (KNN) imputation as a baseline KNN has previously been shown to be a competitive imputation method \citep{bro08}.
   For KNN imputation we first concatenated underlying feature representations from all views to get a joint feature representation. We then sought $k$ nearest data points by using  their available parts, and the missing part was imputed as either average (\knn)\ or the weighted average (\wknn)\ of the selected neighbours. We also compared with an Em-based kernel completion method (\emb)\ proposed by \citet{tsu03}. It cannot solve our problem when no view is complete, hence we study the relative performance only in the cases which it can solve. For \citet{tsu03}'s method we assume the first view is complete. 
   The generative model based method of \citet{lia15} may perform well for data sets with linear kernels but is unlikely to be able to model highly nonlinear kernels. We could not compare with it due to unavailability of code.
   We also compared \emht, \ with  \emhm\ where we assumed the reconstruction weights for all views to be the same , i.e., $\bA^{(m)}=\bA$. 
   
   The hyper-parameters $c_1$ and $c_2$  of \mkc\ and $k$ of \knn\  and \wknn\ were selected with the help of tuning set, from the range of $10^{-3}$ to $10^3$  and $[1,2,3,5,7,10]$ respectively. All reported results indicate performance in the test sets.. 
  

  
\subsection{Prediction error comparisons}


   \subsubsection{Average Relative Error (ARE)}
   We evaluated the performance of all methods using the { \it average relative error} (ARE) \citep{xu13}. Let $\hat{\bk}^{(m)}_{t}$ be the predicted $t^{th}$ row for the $m^{th}$ view and the corresponding true values of kernel row be ${\bk}^{(m)}_{t}$, then the {\it relative error} is the relative root mean square deviation.  
  The average relative error (in percentage) is then computed over all missing data points for a view, that is,
   \begin{equation}
       \textrm{ARE} = \frac{100}{n_{t}^{(m)}}\left(\sum_{t \notin I^{(m)}}\frac{\|\hat{\bk}^{(m)}_{t}-{\bk}^{(m)}_{t}\|_2}{\|{\bk}^{(m)}_{t}\|_2}\right).
       \label{eqn:are}
   \end{equation}
  Here $n_{t}^{(m)}$ is the number of missing samples in the $m^{th}$ view.
   
   \subsubsection{Results}
 
   Table \ref{tab:rmse} shows the Average Relative Error (Equation (\ref{eqn:are})) for the compared methods. It shows that the proposed \mkc\ methods generally predict missing values more accurately than \knn,\ \wknn\ and \emb. In particular, the differences in favor to the \mkc\ methods increase when the number of missing views is increased. The \emb\ sometimes has more than $200\%$ error and  higher (more than $200\%$) variance. The most accurate method in each setup is one of the proposed  \mkc's. \emhm\ is generally the least accurate of them, but still competitive against the other compared methods. We further note that: 
   \begin{itemize}
   \item \emht\ is consistently the best when different views have different kernel functions or eigen-spectra, e.g., TOYLG1  and DREAM (Figure \ref{fig:eigspec}). Better performance of \emht\ than \emhm in DREAM data gives evidence of applicability of \emht in real-world data set. 
   \item \mkckernel\ performs best or very close to \emht\ when kernel functions and eigen-spectra of all views are the same (for instance TOYL, TOYG1 and RCV1/RCV2). As \mkckernel\ learns between-view relationships on kernel values it is not able to perform well for  TOYLG1 and DREAM where very different kernel functions are used in different views. 
   \item \psd\ outperforms all other methods when kernel functions are highly non-linear (such as in TOYG0.1). On less non-linear cases, \psd\ on the other hand trails in accuracy to the other \mkc\ variants. \psd\  is computationally more demanding than the others, to the extent that on RCV1/RCV2 data we had to skip it. 
   \end{itemize}
   \begin{figure}[ht]
       \centering
       \includegraphics[width=0.5\textwidth]{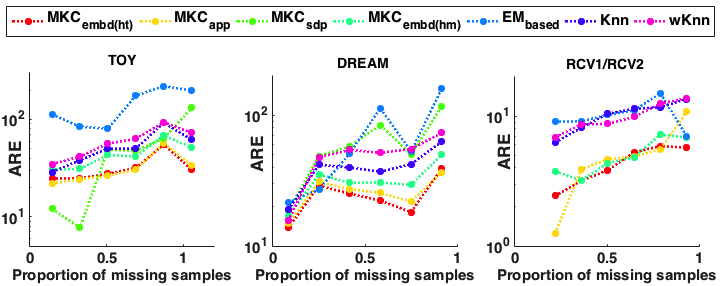}
       \caption{ARE (Equation (\ref{eqn:are})) for different proportions of missing samples. Values are averages over views and random validation and test partitions. The value for TOY are additionally averaged over all 4 TOY data sets. }
       \label{fig:RMSEfrac}
   \end{figure}
   Figure \ref{fig:RMSEfrac} depicts the performance as the number of missing samples per view is increased. Here, \emht,\  \mkckernel\ and \emhm\ prove to be the most robust methods over all data sets. The performance of \psd\ seems to be the most sensitive to amount of missing samples. Overall, \emb, \knn, and \wknn\ have worse error rates than the \mkc\ methods. 
   
   
   \subsection{Running time comparison}
     
   Table \ref{tab:time} depicts the running times for the compared methods. \mkckernel, \emht\ and \emhm\  are many times faster than  \psd.\  In particular, \emhm\ is competitive in running time with the significantly less accurate \emb\ method, except on the RCV1/RCV2 data. As expected, \knn\ and \wknn\ are orders of magnitude faster but fall far from the reconstruction quality of the \mkc\ methods.
   
          \begin{table}[ht]
       \centering
       \scriptsize
       \begin{tabular}{|l|c|c|c|}
      \multicolumn{4}{c}{Number of missing views = 1 (TOY and RCV1/RCV2) and 1 (DREAM)}\\ \hline
      Algorithm & TOY(mins)&DREAM(mins)&	\hspace{-0.3cm}RCV1/RCV2(hrs)	\hspace{-0.3cm} \\ \hline  
 \emht\ \hspace{-0.3cm}&   5.00( $\pm$    2.04) 	\hspace{-0.3cm}&   0.86( $\pm$    0.29) 	\hspace{-0.3cm}&  45.93( $\pm$    2.27) 	\\  
  \mkckernel\ \hspace{-0.3cm}&   2.91( $\pm$    0.39) 	\hspace{-0.3cm}&   1.89( $\pm$    0.62) 	\hspace{-0.3cm}&  16.59( $\pm$    0.28) 	\\  
 \psd\ \hspace{-0.3cm}&  14.82( $\pm$    4.39) 	\hspace{-0.3cm}&   1.13( $\pm$    0.11) 	\hspace{-0.3cm}&   -	\\  
 \emhm\ \hspace{-0.3cm}&   0.15( $\pm$    0.07) 	\hspace{-0.3cm}&   0.05( $\pm$    0.03) 	\hspace{-0.3cm}&   0.28( $\pm$    0.02) 	\\  
 \hline
 \emb\ \hspace{-0.3cm}&   0.50( $\pm$    0.19) 	\hspace{-0.3cm}&   0.03( $\pm$    0.05) 	\hspace{-0.3cm}&   0.03( $\pm$    0.00) 	\\  
\hline
 \multicolumn{4}{c}{Number of missing view = 2 (TOY and RCV1/RCV2) and 3 (DREAM)}\\ \hline
  Algorithm & TOY(mins)&DREAM(mins)&	\hspace{-0.3cm}RCV1/RCV2(hrs)	\hspace{-0.3cm} \\ \hline
\emht\ \hspace{-0.3cm}&   7.58( $\pm$    2.18) 	\hspace{-0.3cm}&   1.13( $\pm$    0.12) 	\hspace{-0.3cm}&  25.86( $\pm$    0.36) 	\\  
 \mkckernel\ \hspace{-0.3cm}&   2.78( $\pm$    0.68) 	\hspace{-0.3cm}&   1.29( $\pm$    0.25) 	\hspace{-0.3cm}&  34.42( $\pm$    1.28) 	\\  
 \psd\ \hspace{-0.3cm}&  25.65( $\pm$    5.43) 	\hspace{-0.3cm}&   1.97( $\pm$    0.34) 	\hspace{-0.3cm}&   -	\\  
 \emhm\ \hspace{-0.3cm}&   0.11( $\pm$    0.05) 	\hspace{-0.3cm}&   0.03( $\pm$    0.01) 	\hspace{-0.3cm}&   0.47( $\pm$    0.02) 	\\  
 \hline
 \emb\ \hspace{-0.3cm}&   0.45( $\pm$    0.08) 	\hspace{-0.3cm}&   0.06( $\pm$    0.06) 	\hspace{-0.3cm}&   0.03( $\pm$    0.00) 	\\  
\hline
 \multicolumn{4}{c}{Number of missing views = 3 (TOY and RCV1/RCV2) and 5 (DREAM)}\\ \hline
   Algorithm & TOY(mins)&DREAM(mins)&	\hspace{-0.3cm}RCV1/RCV2(hrs)	\hspace{-0.3cm} \\ \hline
 \emht\ \hspace{-0.3cm}&   6.83( $\pm$    2.14) 	\hspace{-0.3cm}&   3.39( $\pm$    1.11) 	\hspace{-0.3cm}&  24.39( $\pm$    2.13) 	\\  
 \mkckernel\ \hspace{-0.3cm}&   2.20( $\pm$    0.66) 	\hspace{-0.3cm}&   3.64( $\pm$    1.79) 	\hspace{-0.3cm}&  20.26( $\pm$    1.72) 	\\  
 \psd\ \hspace{-0.3cm}& 178.1( $\pm$  162.9) 	\hspace{-0.3cm}&   4.94( $\pm$    2.48) 	\hspace{-0.3cm}&   -	\\  
 \emhm\ \hspace{-0.3cm}&   0.12( $\pm$    0.08) 	\hspace{-0.3cm}&   0.03( $\pm$    0.02) 	\hspace{-0.3cm}&   0.57( $\pm$    0.00) 	\\  
 \hline
 \emb\ \hspace{-0.3cm}&   0.45( $\pm$    0.05) 	\hspace{-0.3cm}&   0.10( $\pm$    0.05) 	\hspace{-0.3cm}&   0.03( $\pm$    0.00) 	\\  
\hline
 \end{tabular}
 \caption{Average running time over all views and for TOY, over all 4 data sets. The running times for \knn\ and \wknn\ are around  $10^{-3}$ minutes for all data sets.}
 \label{tab:time}
   \end{table}
   

\section{Conclusion}
In this paper,  we have introduced new methods for kernel completion in the multi-view setting. The methods are able to propagate relevant information across views to predict missing rows/columns of kernel matrices in multi-view data. In particular, we are able to predict missing rows/columns of kernel matrices for non-linear kernels, and do not need any complete kernel matrices {\it a priori}.

Our method of within-view learning approximates the full kernel by a sparse basis set of examples with local reconstruction weights, picked up by $\ell_{2,1}$ regularization. This approach has the added benefit of circumventing the need of an explicit PSD constraint in optimization. For learning between views, we proposed two alternative approaches, one based on learning convex kernel combinations and another based on learning a convex set of reconstruction weights. The heterogeneity of the kernels in different views affects which of the approaches is favourable.

Our experiments show that the proposed multi-view completion methods are in general more accurate than previously available methods. In terms of running time, due to the inherent non-convexity of the optimization problems, the new proposals have still room to improve. However, the methods are amenable for efficient parallelization, which we leave for further work.


\bibliography{sahely}
\bibliographystyle{icml2016}
\twocolumn[
\icmltitle{ Multi-view Kernel Completion  Supplementary Material }
]

\section*{Algorithm to solve \psd\ } 
In this section the Algorihtm \ref{algo:psd} describes the algorithm to solve \psd (Equation (\ref{eqn:mkc_sdp})).

\begin{algorithm}
\caption{\hspace{-0.1cm}.  \psd\ $\left(\bK^{(m)},\bI^{(m)}, \forall m \in[1,\ldots,M] \right)$}\label{algo:psd}
\begin{algorithmic}
\STATE Initiaization: 
\STATE $s^0_{mm}=0$, $s^0_{ml}=\frac{1}{M-1}$, \STATE 
\STATE $\hat{\bK}^{(m)^0}_{\bI^{(m)}\bI^{(m)}}= \bK^{(m)}_{\bI^{(m)}\bI^{(m)}}$ and
\STATE $\hat{k}^{(m)^0}_{tt'} \sim uniform(-1,1)$ for all $t,t' \notin \bI^{(m)}$

\vspace{0.5cm}
\REPEAT 
\vspace{0.2cm}
\FOR {m=1 to M } 
\STATE  $ \hat{\bK}^{(m)^k} = \arg\min_{\hat{\bK}^{(m)}} Kobj^k_{\bS}(\hat{\bK}^{(m)})$ 
\STATE \hspace{3cm}         [according to Equations (\ref{eqn:mkc_psd_K})]
\ENDFOR
\vspace{0.2cm}
\FOR{m=1  to M}
\STATE $\bs_m^k = \arg\min_{\bs_m}  Sobj_{[psd]\hat{\bK}^{(m), m=1,\ldots,M}}^k(\bs_m) $ 
\STATE \hspace{3cm}[according to Equation (\ref{eqn:mkc_psd_S})]
\ENDFOR
\vspace{0.2cm}
\UNTIL{converge}
\end{algorithmic}
\end{algorithm}

The Equation (\ref{eqn:mkc_sdp}) has two sets of unknowns, $\bS$ and the $\hat{\bK}^{(m)}$'s. We update $\hat{\bK}^{(m)}$ and $\bS$  in an iterative manner. In the $k^{th}$ iteration, for fixed a $\bS^{k-1}$ and $\hat{\bK}^{(m)^{k-1}}$, the $\hat{\bK}^{(m)}$ is updated by independently by solving following Semi-definite Programming:
\begin{eqnarray}
\hat{\bK}^{(m)^{k}} &=& \arg\min_{\bK}  Kobj_{\bS}^k(\hat{\bK}^{(m)}) \nonumber \\
s.t && \hat{\bK}^{(m)} \succeq 0,
\label{eqn:mkc_psd_K}
\end{eqnarray}
where 
 $Kobj_{\bS}^k(\hat{\bK}^{(m)}) = \|\hat{\bK}^{(m)}_{\bI^{(m)}\bI^{(m)}} -{\bK}^{(m)}_{\bI^{(m)}\bI^{(m)}} \|_2^2 + c \|\hat{\bK}^{(m)}- \sum_{l=1, l\ne m}^{M}s^{k-1}_{ml}\hat{\bK}^{(l)^{k-1}} \|_2^2 + c \sum_{l=1, l\ne m}^{M} \|\hat{\bK}^{(l)^{k-1}}- \sum_{l'=1, l' \ne l,m}^{M} s^{k-1}_{ll'} \hat{\bK}^{(l')^{k-1}} - s^{k-1}_{lm}\hat{\bK}^{(m)}\|_2^2$

Again, in the $k^{th}$ iteration, for fixed $\hat{\bk}^{(m)^k}, \forall m=[1,\ldots,M]$,  $\bS$ is updated by  independently updating each row ($\bs_m$) through solving the  following Quadratic Program:
\begin{eqnarray}
\bs_m^{k} &=& \arg\min_{\bs_m}  Sobj_{[psd]\hat{\bK}^{(m), m=1,\ldots,M}}^k(\bs_m) \nonumber \\
s.t&& \sum_{l\ne m} s_{ml} =1, \nonumber \\
&& s_{ml} \ge 0 \,\, \forall l
\label{eqn:mkc_psd_S}
\end{eqnarray}

Here 
 $ Sobj_{[psd]\hat{\bK}^{(m), m=1,\ldots,M}}^k(\bs_m) = \|\hat{\bK}^{(m)^k}- \sum_{l=1, l\ne m}^{M}s_{ml}\hat{\bK}^{(l)^k} \|_2^2  $.

\section*{Algorithm to solve \mkckernel\ }

In this section the Algorithm \ref{algo:app} describes the main algorithm to solve \mkckernel (Equation (\ref{eqn:mkc_app})).

\begin{algorithm}
\caption{\hspace{-0.1cm}.  \mkckernel \begin{small}$\left(\bK^{(m)},\bI^{(m)}, \forall m \in[1,\ldots,M] \right)$ \end{small}}
\label{algo:app}
\begin{algorithmic}
\STATE Initiaization: 
\STATE $s^0_{mm}=0$, $s^0_{ml}=\frac{1}{M-1}$, 
\STATE $\bA^{(m)^0}_{\bI^{(m)}\bI^{(m)}}$ = Identity matrix and
\STATE $a^{(m)^0}_{tt'} \sim uniform(-1,1)$ for all $t,t' \notin \bI^{(m)}$ 
\vspace{0.5cm}
\REPEAT 
\FOR{m=1 to M } 
\FOR {t=1  to N}
\STATE  $ \ba_t^{(m)^k} = max\left(0, (1-\frac{\lambda c_2}{\|\Delta \ba
    _t^{(m)^{k-1}}\|_2}) \Delta \ba_t^{(m)^{k-1}}\right)$
\STATE \hspace{2cm}[according to Equations (\ref{eqn:prox_app}, \ref{eqn:prox_m_app})]
\STATE \hspace{3cm} [$\lambda$ is fixed by line search]
\ENDFOR
\ENDFOR
\FOR{m=1  to M}
\STATE $\bs_m^k = \arg\min_{\bs_m}  Sobj_{[app]\bA}^k(\bs_m) $
\STATE \hspace{3cm}[according to Equation (\ref{eqn:mkc_app_S})]\ENDFOR
\UNTIL{converge}
\end{algorithmic}
\end{algorithm}

Substituting $\hat{\bK}^{(m)} =  \bA^{(m)^T}_{\bI^{(m)}} \bK^{(m)}_{I^{(m)},I^{(m)}} \bA^{(m)}_{\bI^{(m)}}$, the Equation (\ref{eqn:mkc_app}) has two sets of unknowns, $\bS$ and the $\bA^{(m)}$'s. We update  $\bA^{(m)}$ and $\bS$ in an iterative manner. In the $k^{th}$ iteration for a fixed $\bS^{k-1}$ from previous iteration, to update $\bA^{(m)}$'s  we need to solve following for each  $m$:

\begin{eqnarray}
 \bA^{(m)^k} = \arg\min_{\bA^{(m)}}   Aobj_{[app]\bS}^k(\bA^{(m)}) +c_2 \Omega(\bA^{(m)^{k-1}}) \nonumber
\end{eqnarray}
\label{eqn:mkc_app2}

where $ \Omega(\bA^{(m)})=\|\bA^{(m)}\|_{2,1}$  and  $Aobj_{[app]\bS}^k(\bA^{(m)})=\|\bK^{(m)}_{\bI^{(m)}\bI^{(m)}}- \left[\bA^{(m)^T}_{\bI^{(m)}}\bK^{(m)}_{\bI^{(m)}\bI^{(m)}}\bA^{(m)}_{\bI^{(m)}}\right]_{\bI^{(m)}\bI^{(m)}}\|_2^2 +c_1 \sum_{m=1}^M  \|\bA^{(m)^T}_{\bI^{(m)}}\bK^{(m)}_{\bI^{(m)}\bI^{(m)}}\bA^{(m)}_{\bI^{(m)}}- \sum_{l=1, l\ne m}^{M}s^{k-1}_{ml}\bA^{(l)^T}_{\bI^{(l)}}\bK^{(l)}_{\bI^{(l)}\bI^{(l)}}\bA^{(l)}_{\bI^{(l)}}  \|_2^2$.

Instead of solving this problem in each iteration we update $\bA^{(m)}$ using proximal gradient descent. Hence, in each iteration,
\begin{small}
\begin{eqnarray}
    \bA^{(m)^k} = Prox_{\lambda c_2\Omega}\left(\bA^{(m)^{k-1}}-\lambda \partial Aobj_{[app]\bS}^k(\bA^{(m)^{k-1}}) \right)
  \label{eqn:prox_app}
\end{eqnarray}
\end{small}

where $\partial Aobj_{[app]\bS}^k(\bA^{(m)^{k-1}})$ is the differential of $Aobj_{[app]\bS}^{k}(\bA^{(m)})$ at $\bA^{(m)^{k-1}}$ and $\lambda$ is the step size which is decided by a line search. 
By applying proximal operator on each row of $\bA$ (i.e., $\ba_{t}$)  in Equation (\ref{eqn:prox_app}) the  solution of Equation (\ref{eqn:prox_app}) is 

\begin{equation}
    \ba_t^{(m)^k} = max\left(0, (1-\frac{\lambda c_2}{\|\Delta \ba
    _t^{(m)^{k-1}}\|_2}) \Delta \ba_t^{(m)^{k-1}}\right),
\label{eqn:prox_m_app}
\end{equation}

where $\Delta \ba_t^{(m)^{k-1}}$ is  the $t^{th}$ row of  $\left(\bA^{(m)^{k-1}}-\lambda \partial Aobj_{[app]\bS}^k(\bA^{(m)})\right)$.

Again, in the $k^{th}$ iteration, for fixed $\bA^{(m)^k}, \forall m=[1,\ldots,M]$,  $\bS$ is updated by  independently updating each row ($\bs_m$) through solving the  following Quadratic Program:
\begin{small}
\begin{eqnarray}
\bs_m^{k} &=& \arg\min_{\bs_m}  Sobj_{[app]\bA}^k(\bs_m) \nonumber \\
s.t&& \sum_{l\ne m} s_{ml} =1, \nonumber \\
&& s_{ml} \ge 0 \,\, \forall l
\label{eqn:mkc_app_S}
\end{eqnarray}

\end{small}

where 
$  Sobj_{[app]\bA}^k(\bs_m) = \|\bA^{(m)^T}_{\bI^{(m)}}\bK^{(m)}_{\bI^{(m)}\bI^{(m)}}\bA^{(m)}_{\bI^{(m)}}- \sum_{l=1, l\ne m}^{M}s_{ml}\bA^{(l)^T}_{\bI^{(l)}}\bK^{(l)}_{\bI^{(l)}\bI^{(l)}}\bA^{(l)}_{\bI^{(l)}}  \|_2^2  $.

\end{document}